%% file: main.tex
\documentclass[letterpaper, 10 pt, conference]{ieeeconf}  % Comment this line out if you need a4paper

\IEEEoverridecommandlockouts                              % This command is only needed if 
\usepackage{graphics}
\usepackage{amsmath}
\usepackage{amssymb}
\usepackage{xcolor}
\usepackage{bbold}
\usepackage{bm}
\usepackage{optidef}
\usepackage{mathrsfs}  % script font
\usepackage[noadjust]{cite}
\usepackage{lipsum}
\usepackage{todonotes}
\usepackage{xfrac}  % for \sfrac
\usepackage{bigdelim}
\usepackage{tikz}
\usetikzlibrary{backgrounds, chains}
\usetikzlibrary{arrows.meta}
\newcommand{\comment}[1]{} % hide comments

\definecolor{patrick_color}{rgb}{.0,.6,.05}
\definecolor{gerry_color}{rgb}{.5,.7,.1}
\definecolor{frank_color}{rgb}{0.75,0.25,0.0}
\definecolor{seth_color}{rgb}{0.6,0.0,0.6}

\makeatletter
\let\NAT@parse\undefined
\makeatother
\usepackage{hyperref}
\hypersetup{
    pdftitle={Generalizing the Trajectory Retiming Problem to Quadratic Objective Functions},
    pdfauthor={Gerry Chen and Frank Dellaert and Seth Hutchinson},
    pdfsubject={Robotic Trajectory Retiming},
    pdfkeywords={trajectory retiming, time-optimal path parameterization, TOPP, TOPP-RA, factor graph, variable elimination},
    bookmarksnumbered=true,     
    bookmarksopen=true,         
    bookmarksopenlevel=1,       
    colorlinks=true,
    allcolors=blue,
    pdfstartview=Fit,           
    pdfpagemode=UseOutlines,
    }
\usepackage{hypcap} % link figures to image not caption

\input{macros}

\title{\LARGE \bf
Generalizing Trajectory Retiming to Quadratic Objective Functions
}

\author{Gerry Chen, Frank Dellaert, and Seth Hutchinson\vspace*{-0.15em}%
\thanks{This work was supported by NSF Grant No. 2008302.}%
\thanks{All authors are with 
the Institute for Robotics and Intelligent Machines (IRIM), 
Georgia Institute of Technology, Atlanta, GA 30332, USA
        {\tt\small \{gchen,fd27,seth\}@gatech.edu}}%
}

\begin{document}

\maketitle
\thispagestyle{empty}
\pagestyle{empty}

%%%%%%%%%%%%%%%%%%%%%%%%%%%%%%%%%%%%%%%%%%%%%%%%%%%%%%%%%%%%%%%%%%%%%%%%%%%%%%%%
\begin{abstract}

% Trajectory optimization is a fundamental problem in robotics in which kino-dynamic, obstacle avoidance, and/or other constraints must be satisfied while optimizing some objective function.
Trajectory retiming is the task of computing a feasible time parameterization to traverse a path.  It is commonly used in the decoupled approach to trajectory optimization whereby a path is first found, then a retiming algorithm computes a speed profile that satisfies kino-dynamic and other constraints.
% We focus on the decoupled approach to trajectory optimization whereby first a path is found, then a retiming algorithm computes a trajectory.
% Trajectory optimization can be solved either as one optimization problem or as decoupled path-planning and retiming problems.
While trajectory retiming is most often formulated with the minimum-time objective (\ie traverse the path as fast as possible), it is not always the most desirable objective, particularly when we seek to balance multiple objectives or when bang-bang control is unsuitable.
% in applications where we seek to balance multiple objectives or where matching a desired speed, torque, or other profile is preferable over one which rides along constraint boundaries to minimize time.
In this paper, we present a novel algorithm based on factor graph variable elimination that can solve for the global optimum of the retiming problem with \emph{quadratic} objectives as well (\eg minimize control effort or match a nominal speed by minimizing squared error)%
, which may extend to arbitrary objectives with iteration%
. % or the standard minimal-time objective.
% Retiming with quadratic objectives has applications in settings where multiple objectives should be balanced, or where matching a desired speed, torque, or other profile is preferable over one which rides along constraint boundaries to minimize time.
% Our algorithm extends prior works in time-optimal trajectory retiming by
% % re-interpreting the previous reachability analysis approach as a factor graph variable elimination algorithm which permits
% being able to solve the trajectory retiming problem quadratic objectives in addition to the standard minimal-time objective.
Our work extends prior works, which find only solutions on the boundary of the feasible region, while maintaining the same linear time complexity from a single forward-backward pass.
% Like prior works, we apply a single forward and backward pass to compute the optimal retiming in linear time complexity.  However, unlike many previous works which leverage the monotonicity of the minimal-time objective to assume a bang-bang solution, our algorithm allows for also solving the wider set of quadratic objectives which may have solutions that do not lie on the boundary of the feasible region.
We experimentally demonstrate that (1) we achieve better real-world robot performance by using quadratic objectives in place of the minimum-time objective, and (2) our implementation is comparable or faster than state-of-the-art retiming algorithms.
% We demonstrate the effectiveness of our algorithm with an example cable-driven robot path tracking problem.
\vspace{-.4em}
\end{abstract}
%
%%%%%%%%%%%%%%%%%%%%%%%%%%%%%%%%%%%%%%%%%%%%%%%%%%%%%%%%%%%%%%%%%%%%%%%%%%%%%%%%
\section{Introduction} \label{sec:intro}
\input{_intro}

%%%%%%%%%%%%%
\section{Approach}
In this section, we start by providing a brief recap of the standard reparameterizations used in trajectory retiming.  We then introduce factor graphs by explaining the TOPP-RA (\emph{R}eachability \emph{A}nalysis) algorithm using a factor graph interpretation.  Finally, we use the factor graph approach to extend the TOPP-RA algorithm to the QOPP problem.

\subsection{Reparameterization} \label{ssec:reparam}
\input{_approach_reparam}

\subsection{Solving the TOPP Problem with Factor Graphs} \label{ssec:topp}
\input{_approach_topp}

\subsection{Solving the QOPP Problem with Factor Graphs} \label{ssec:quadratic_approach}
\input{_approach_quadratic}

%%%%%%%%%%%%%
\section{Experimental Results} \label{sec:results}
\input{_results}

%%%%%%%%%%%%%
\section{Discussion} \label{sec:discussion}
\input{_discussion}

%%%%%%%%%%%%%
\section{Conclusions and Future Works} \label{sec:conclusions}
\input{_conclusions}

%%%%%%%%%%%%%%%%%%%%%%%%%%%%%%%%%%%%%%%%%%%%%%%%%%%%%%%%%%%%%%%%%%%%%%%%%%%%%%%%

\addtolength{\textheight}{-9.1cm}   % This command serves to balance the column lengths

%%%%%%%%%%%%%%%%%%%%%%%%%%%%%%%%%%%%%%%%%%%%%%%%%%%%%%%%%%%%%%%%%%%%%%%%%%%%%%%%
\section*{APPENDIX}

In this section we provide a brief introduction to factor graph variable elimination for optimal control.
\label{sec:app:factor_graphs}
\input{_appendix_fg_background}

% \section*{ACKNOWLEDGMENT}

% aoeu

%%%%%%%%%%%%%%%%%%%%%%%%%%%%%%%%%%%%%%%%%%%%%%%%%%%%%%%%%%%%%%%%%%%%%%%%%%%%%%%%
\clearpage

\bibliographystyle{IEEEtran}
\bibliography{IEEEabrv,topp,common,mine}
% \printbibliography

\end{document}

%% file: macros.tex
\newcommand{\vect}{\bm{t}}

\newcommand{\vecx}{\bm{x}}

\newcommand{\vecq}{\bm{q}}

\newcommand{\vecqd}{\dot{\vecq}}
\newcommand{\vecxd}{\dot{\vecx}}
\newcommand{\vecqdd}{\ddot{\vecq}}

\newcommand{\boldA}{\bm{A}}
\newcommand{\boldB}{\bm{B}}
\newcommand{\boldf}{\bm{f}}

\newcommand{\boldW}{\bm{W}}
\newcommand{\boldC}{\bm{C}}

\newcommand{\wrench}{\bm{F}}
\newcommand{\norm}[1]{\left\lVert#1\right\rVert}
\newcommand{\ff}{{f\kern-0.15em f}} % feedforward
\newcommand{\deriv}[2]{\frac{\mathrm{d}#1}{\mathrm{d}#2}}
\newcommand{\dderiv}[2]{\frac{\mathrm{d}^2#1}{\mathrm{d}#2^2}}

\newcommand{\deltas}{\Delta_s}

\newcommand{\xt}{{\tilde{x}}}
\newcommand{\ut}{{\tilde{u}}}

\usepackage{xspace}

\newcommand*{\eg}{e.g.\@\xspace}
\newcommand*{\ie}{i.e.\@\xspace}

\makeatletter
\newcommand*{\etc}{%
    \@ifnextchar{.}%
        {etc}%
        {etc.\@\xspace}%
}
\makeatother

%% file: _intro.tex
Trajectory optimization is necessary to execute safe, effective robot motions 
% and is often formulated as a constrained nonlinear optimization problem which 
and can be solved either directly as a state-space optimization problem or with a decoupled approach in which a configuration-space path is first found then retimed \cite[Ch. 11.2]{Choset05book_PrinciplesRobotMotion}.
% Although several approaches can successfully solve the large state-space optimization problems, they are often computationally expensive and may lack formal guarantees on optimality without additional assumptions, especially for high-dimensional, complex systems.
We focus on the decoupled approach to separate the trajectory optimization problem into a path planning problem and a retiming problem.  In this approach, a path planner first generates a path which \eg avoids obstacles and solves the robot kinematics.  Then, a retiming algorithm is applied to the path to generate a trajectory which satisfies the robot's dynamics and other constraints.  Although the resulting solution is only an approximation to the original trajectory optimization problem, it is often more tractable, faster, and reliable, especially when the path planner has good heuristics for kino-dynamic constraints or the task specification defines the path.

The \emph{time-optimal} trajectory retiming problem (also known as time-optimal path parameterization (TOPP), time-optimal path tracking, and several other names), in which the objective is to minimize the time to traverse the path, represents the most common retiming objective.  This problem has been studied extensively in the literature and traditionally has taken one of three approaches \cite{Nagy19tro_Peaked_SequentialTimeOptimalPathTracking}: convex optimization, dynamic programming, and searching for bang-bang control switching points where the active constraints change.  % Of course, the approaches share many similarities as the latter two are solving the same convex optimization problem, but just leveraging its special structure to solve it more efficiently. 

Two key precursor techniques enabling efficient algorithms are shared by almost all approaches, dating back to at least \cite{Bobrow85ijrr_TimeOptimalControlRobotic,Shin85tac_MinimumtimeControlRobotic,Pfeiffer87jra_ConceptManipulatorTrajectory
,% } with so-called ``zero-inertia points'' handled by \cite{
Slotine89tro_ImprovingEfficiencyTimeoptimal,Singh87jdsme_OptimalTrajectoryGeneration}.
First, the equations of motions and constraints are reparameterized in terms of the \emph{scalar} time parameterization function (see Sec. \ref{ssec:reparam} for details).  This enables Bellman-style forward-backward algorithms.
% is re-parameterized with the path acceleration and the square of the speed to convert the problem into an optimization over scalar functions with linear equality and inequality constraints and an inverse square-root objective.  
% The problem is discretized to become a convex optimization problem: one with linear equality and inequality constraints and a convex objective function.
% For typical equations of motion and constraints, the dynamics and inequalities are actually linear in the acceleration and speed squared making it amenable to optimization routines.
Second, the fact that the time-optimal objective is monotonic implies that the solution must lie on the boundary of the feasible set (bang-bang).  Hauser \cite{Hauser13rss_FastInterpolationTimeOptimization}, Nagy \& Vajk \cite{Nagy19tro_Peaked_SequentialTimeOptimalPathTracking}, and Pham et. al. \cite{Nagy19tro_Peaked_SequentialTimeOptimalPathTracking} all utilize proofs along these lines to justify the use of sequential linear programming (SLP) or greedy speed maximization during the backward pass. %\cite{Nagy19tro_Peaked_SequentialTimeOptimalPathTracking,Pham18tro_NewApproachTimeOptimal}.
% Hauser provides a beautifully succinct proof of this fact, which is used to show that the problem can be solved using sequential linear programming (SLP) \cite{Hauser13rss_FastInterpolationTimeOptimization}.  Nagy and Vajk go a step further using ``peaked constraint'' terminology to show that the objective function (over a vector variable) can be replaced by any strictly monotone decreasing function such as $-\mathbb{1}^T \bm{x}$ \cite{Nagy19tro_Peaked_SequentialTimeOptimalPathTracking}, which essentially generalizes the proof in \cite{Pham18tro_NewApproachTimeOptimal} that a greedy speed maximization strategy is optimal.
% With a few exceptions discussed in the next paragraph, most works in retiming literature utilize t
This bang-bang approach is efficient but restricts the objective to minimum-time or similar objectives.

\begin{figure}
  \centering \vspace*{-0.3em}
  \includegraphics[width=3.2in]{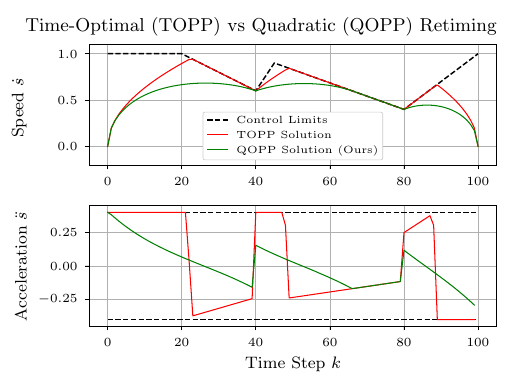}\vspace*{-0.6em}
  \caption{Using quadratic objectives (green) in place of the minimum-time objective (red) results in less aggressive maneuvers and a more consistent speed profile.  Our approach may increase safety margins, improve tracking accuracy, and reduce premature wear by balancing objectives like execution speed and motor torques.}
  \label{fig:topp_vs_qopp}
\end{figure}
\begin{figure}
  \centering
  \includegraphics*[width=\linewidth]{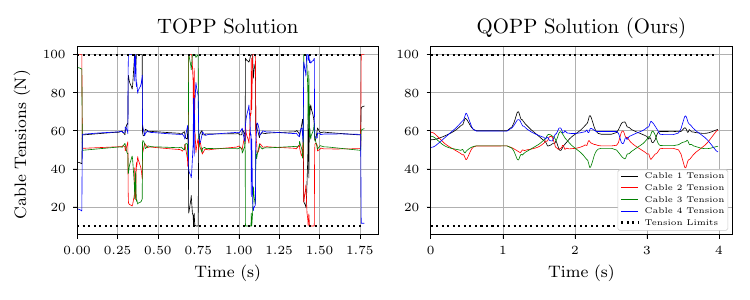}\vspace*{-0.6em}
  \caption{Open-loop cable robot tensions for the TOPP trajectory frequently hit control limits, losing stiffness and stability.  Meanwhile, quadratic objectives (QOPP) enable a tunable tradeoff between duration and safety margin.}
  \label{fig:tensions_compare}
\end{figure}

% Perhaps the most noteworthy advancements came from TOPP-RA \cite{Pham18tro_NewApproachTimeOptimal} who recognized that the LP can be solved with a \emph{R}eachability \emph{A}nalysis approach, which is essentially a symbolic dynamic programming algorithm for propagating constraints.  This approach is particularly elegant as it is both intuitive and only requires a single forward and backward pass to compute the optimal retiming in linear time.  As we will see in Section \ref{ssec:topp}, this approach can also be interpreted using the factor graph variable elimination algorithm.

However, the \emph{time-optimal} retiming problem is not always the most desirable objective \cite{Verscheure09tac_TimeOptimalPathTracking}, particularly in applications where we seek to balance multiple objectives or where bang-bang control is unsuitable.
For example, cable-driven parallel robots maintain stiffness primarily through internal tension which diminishes the closer they are to torque limits.  Thus balancing speed with torque margin is desirable to maintain stability and safety \cite{Chen22iros_LocallyOptimalEstimation,Chen22icra_GTGraffitiSprayPainting}.
% Painting applications seek to maintain a consistent speed to ensure a uniform paint thickness but should allow this secondary objective to be relaxed when actuator limits are reached \cite{Chen22icra_GTGraffitiSprayPainting}.
This and several other applications in balancing robot safety, stability, and wear with speed of operation motivate the use of quadratic objectives, which can minimize the sum of multiple squared errors instead of or in addition to hard constraints with TOPP.

Surprisingly alternate retiming objectives are rarely considered in the literature.  Dynamic programming approaches \cite{Shin86tac_DynamicProgrammingApproach,Pfeiffer87jra_ConceptManipulatorTrajectory,Singh87jdsme_OptimalTrajectoryGeneration} discretize not only in time but also in state space.  Some approaches address other objectives, especially energy-minimization \cite{Constantinescu00jrs_SmoothTimeoptimalTrajectory,Shiller94Proceedingsofthe1994IEEEInternationalConferenceonRoboticsandAutomation_TimeenergyOptimalControl}, but apply only to specific objectives, \eg integral of a time-independent running cost.  Direct transcription approaches tend to be the most general \cite{Betts93jgcd_PathconstrainedTrajectoryOptimization,Verscheure09tac_TimeOptimalPathTracking} but, even with second-order cone problem or sparse linear algebra solvers, do not fully exploit the structure of this scalar-function optimal control problem.

In this paper, we present a novel, linear-time algorithm for solving \emph{Quadratic} Objective Path Parameterization (QOPP).  Our algorithm matches the performance of TOPP-RA \cite{Pham18tro_NewApproachTimeOptimal} on TOPP problems while also solving QOPP problems in linear time.
Our contributions are as follows:
\begin{enumerate}
  \item We share a \textbf{re-interpretation} of the TOPP-RA \cite{Pham18tro_NewApproachTimeOptimal} approach as factor graph variable elimination.
  \item We \textbf{extend} the TOPP-RA approach to solve the trajectory retiming problem with \textbf{quadratic objectives}.
  \item We describe how the quadratic case can be extended to arbitrary \textbf{nonlinear objectives} with iteration.
  \item We write and benchmark a fast, C++ \textbf{implementation}.
  \item We validate the utility of objectives other than minimum time with real-world robot \textbf{experiments}.
\end{enumerate}

% Fourier-Motzkin elimination

%% file: _approach_reparam.tex
% We define our TOPP problem drawing from the notation and work of \cite{Pham18tro_NewApproachTimeOptimal} as follows.
We define the TOPP problem as follows.
Given a path $\vecq(s): [0, 1]\rightarrow \mathbb{R}^n$, we seek to find a reparameterization, $s(t)$ is a monotonically increasing function from $[0, T]\rightarrow[0, 1]$, which minimizes the total time $T$ while satisfying a set of general first and second order constraints:
\begin{argmini!}|l|[1]
  {s(t)}{T\label{eq:topp_objective}}{\label{eq:topp_problem}}{s^*(t) =}
  % \addConstraint{\boldA^{\vecq}(s)\vecqdd + \vecqd^T\boldB^{\vecq}(s)\vecqd + \boldf^{\vecq}(s)}{\in \mathscr{C}^{\vecq}(s) \label{eq:topp_problem_joint2}}
  % \addConstraint{\boldA^{\vecqd}(s)\vecqd + \boldf^{\vecqd}(s)}{\in \mathscr{C}^{\vecqd}(s) \label{eq:topp_problem_joint1}}
  \addConstraint{\boldA(s)\vecqdd + \vecqd^T\boldB(s)\vecqd + \boldf(s)}{\in \mathscr{C}(s) \label{eq:topp_problem_task2}}
  \addConstraint{\boldA^{v}(s)\vecqd + \boldf^{v}(s)}{\in \mathscr{C}^{v}(s) \label{eq:topp_problem_task1}}
\end{argmini!}
where 
$\boldA, \boldB, \boldf$ denote coefficient matrix-, tensor-, and vector- valued functions for general second order constraints; % TODO: fix underfull spacing
$\boldA^v, \boldf^v$ denote coefficient matrix- and vector- valued functions for general first-order constraints;
and $\mathscr{C}, \mathscr{C}^v$ denote convex polytope-valued functions of admissible values for the corresponding constraints.
For notational convenience, we will assume these to be closed polytopes (so we can use $\le$), but everything applies to open intervals as well.
Arguments for $\vecq(s(t)), \vecqd(s(t)), \vecqdd(s(t))$, and $s(t)$ in \eqref{eq:topp_problem_task2} and \eqref{eq:topp_problem_task1} were omitted for readability.  Equations \eqref{eq:topp_problem_task2} and \eqref{eq:topp_problem_task1} apply to $t \in [0, T]$ (omitted for readability).
Although rigid-body equations of motion are most commonly notated with $\boldC(\vecq, \vecqd)\vecqd$ instead of $\vecqd^T\boldB(\vecq)\vecqd$, the latter is also standard \cite{Lynch17book_ModernRobotics,Choset05book_PrinciplesRobotMotion} and valid for conservative systems.

% \subsection{TOPP Solution}
% We solve the TOPP problem by applying the same method outlined in \cite{Pham18tro_NewApproachTimeOptimal}.  This relies on 3 key insights:
% \begin{enumerate}
%   \item The task-space constraints in \eqref{eq:topp_problem} can be re-written in terms of $s(t)$ and its derivatives.
%   \item The clever parameterizations $x=\dot{s}^2$ and $u=\ddot{s}$ allows us to rewrite the $s(t)$ constraints as a scalar linear dynamical system with linear control-action constraints.
%   \item The linear dynamical system forms a linear program which can be solved in linear time using a dynamic programming algorithm, which is equivalent to variable elimination.
% \end{enumerate}
% We now describe each of these steps.

\subsubsection{Re-writing Constraints in Terms of \texorpdfstring{$s(t)$}{s(t)}}
Differentiating $\vecq(s(t))$ with respect to $t$ yields
\begin{align}
  \vecqd(s(t)) &= \deriv{\vecq}{s} \dot{s} \\
  \ddot{\vecq}(s(t)) &= \dderiv{\vecq}{s} \dot{s}^2 + \deriv{\vecq}{s} \ddot{s}.
\end{align}
Substituting into \eqref{eq:topp_problem_task2} and \eqref{eq:topp_problem_task1} yields constraints of the form:
\begin{align}
  \bm{a}(s)\ddot{s} + \bm{b}(s)\dot{s}^2 + \bm{c}(s) &\in \mathscr{C}(s) \label{eq:topp_problem_task2_st} \\
  \bm{a}^v(s)\dot{s} + \bm{c}^v(s) &\in \mathscr{C}^v(s) \label{eq:topp_problem_task1_st}
\end{align}
% \begin{align}
%   \boldA^{\vecx}(s(t))\left(\dderiv{\vecx}{s} \dot{s}^2 + \deriv{\vecx}{s} \ddot{s}\right) + \nonumber \\
%   \dot{\vecx}^T\boldB^{\vecx}(s(t))\dot{\vecx} + \boldf^{\vecx}(s(t)) &\in \mathscr{C}^{\vecx}(s(t)) \label{eq:topp_problem_task2_st} \\
%   \boldA^{\vecxd}(s(t))\left(\deriv{\vecx}{s} \dot{s}\right) + \boldf^{\vecxd}(s(t)) &\in \mathscr{C}^{\vecxd}(s(t)) \label{eq:topp_problem_task1_st}
% \end{align}
where $\bm{a}, \bm{b}, \bm{c}, \bm{a}^v, \bm{c}^v$ are vector functions of $s$.

\eqref{eq:topp_problem_task1_st} can always be rewritten in the form of \eqref{eq:topp_problem_task2_st} since $\dot{s}$ is a scalar, positive function.  Thus \eqref{eq:topp_problem_task1_st} can be written
$\dot{s}_{min}\le\dot{s}\le \dot{s}_{max}$
% $\dot{s}\in[\dot{s}_{min}, \dot{s}_{max}]$
and squared.  % Specifically, for any particular value of $s$, \eqref{eq:topp_problem_task1_st} is the intersection of a ray with a convex region, which in turn can be reduced to just a convex set in $\mathbb{R}^1$ (\ie $a \le \dot{s} \le b$ if closed).  Furthermore, because $s(t)$ is monotonically increasing, we have another constraint that $\dot{s}(t) > 0$ so we can safely square the constraint.  For example, if \eqref{eq:topp_problem_task1_st} reduces to $\dot{s} \in (a, b)$, then we can write $0\ddot{s} + 1\dot{s}^2 + 0\in (\min (0, a)^2, \min(0, b)^2)$.
% Therefore, we have turned our TOPP problem into:
Our TOPP problem is now:
\begin{argmini!}|l|
  {s(t)}{T}{}{s^*(t) =}
  \addConstraint{\bm{a}(s)\ddot{s} + \bm{b}(s)\dot{s}^2 + \bm{c}(s)}{\in \mathscr{C}(s) \label{eq:topp_problem_constraint}}
\end{argmini!}

\subsubsection{Discretization}
As is common for TOPP convex optimization and dynamic programming approaches (but not all switching-search approaches), we discretize in $s$:
% At this point, to remove the dependence of the coefficients on $s$, we discretize our problem in $s$:
\begin{align}
  a_k\ddot{s}_k + b_k\dot{s}_k^2 + c_k &\in \mathscr{C}_k \label{eq:topp_problem_constraint_discrete}
\end{align}

Defining:\vspace*{-0.75\baselineskip}
\begin{align}
  x := \dot{s}^2,\quad u := \ddot{s},
\end{align}
we can rewrite \eqref{eq:topp_problem_constraint} as
\begin{align}
  \bm{a}_ku_k + \bm{b}_kx_k + \bm{c}_k &\in \mathscr{C}_k \label{eq:topp_problem_constraint_xu}
\end{align}
and, remarkably, $\deriv{x}{s} = 2\dot{s}\deriv{\dot{s}}{s} = 2\deriv{\dot{s}}{s}\deriv{s}{t} = 2\ddot{s}=2u$.  Then assuming a zero-order hold on $u$ (piecewise constant between intervals), we introduce another constraint from the relationship between $x$ and $u$:
\begin{align}
  x_{i+1} = x_k + 2u_k \deltas.
\end{align}

The minimum-time objective can also be discretized:
\begin{align}
  T = \int_0^1 \frac{1}{\dot{s}(t)} ds
  % = \sum_{k=0}^{N-1} \frac{2\deltas}{\dot{s}_k+\dot{s}_{k+1}}
  = \sum_{k=0}^{N-1} \frac{2\deltas}{\sqrt{x_k}+\sqrt{x_{k+1}}},
  % \approx \sum_{k=0}^{N-1} \frac{\deltas}{\dot{s}_k}
\end{align}
where the summation holds exactly for the piecewise-constant assumption on $u$ \cite[Sec 6.1.1]{Lipp14ijc_MinimumtimeSpeedOptimisation}.

Finally, the objective $\min \sum T$ can be achieved by greedily selecting the maximum $\dot{s}_k$ from the set of reachable values at each time step, as long as $\deltas$ is sufficiently small \cite{Pham18tro_NewApproachTimeOptimal}.  Notationally, we can express this ``greedy'' selection optimization problem as:

\begin{maxi!}|l|[1]
  {\substack{x_0, \ldots, x_N,\\u_0, \ldots, u_{N-1}}}{\sum_{i=0}^N w^k x_k \label{eq:topp_objective_final}}{\label{eq:topp_problem_final}}{}
  \addConstraint{\bm{a}_ku_k + \bm{b}_kx_k + \bm{c}_k}{\in \mathscr{C}_k \label{eq:topp_problem_constraint_discr}}{,\quad i=0,\ldots,N}
  \addConstraint{x_{i+1}-x_k-2u_k\deltas}{= 0 \label{eq:topp_problem_dynamics}}{,\quad i=0,\ldots, N-1}
  \addConstraint{x_k}{> 0}{,\quad i=0,\ldots, N. \label{eq:topp_problem_x_positive}}
\end{maxi!}
for some very large $w$, which denotes that each $x_k$ should be taken greedily and irrespective of any other $x_k$.  Define $u_N := 0$ for notational simplicity of \eqref{eq:topp_problem_constraint_discr}.

Intuitively, the equivalence of the optimization problems despite a different objective function is due to the facts that
(a) if a larger $x_k$ never sacrifices $x_{k+1}$ to be smaller, then the greedy approach works, and (b) there exists some critical threshold $\deltas^*$ such that, when $\deltas \le \deltas^*$, the former is true for all $k$.
Specifically, the former is true when the dynamics coefficient $2\deltas$ is smaller than $\bm{a}_k/\bm{b}_k$ for all $k$.
% so that it will never be better to sacrifice a smaller $x_k$ to gain a larger $u_k$ (which in turn gains a larger $x_{i+1}$).
Though the details of the proof require additional machinery to address the fact that $\bm{a}_k / \bm{b}_k$ is not defined for vectors, the intuition is the same and given in \cite{Pham18tro_NewApproachTimeOptimal}.

Although we could remove $u_k$ from the optimization problem \eqref{eq:topp_problem_final} by substituting $u_k = (x_{i+1}-x_k)/2\deltas$ from \eqref{eq:topp_problem_dynamics}, we opt not to because (1) it will make min-effort objectives more natural when we extend to quadratic objectives and (2) the factor graph elimination will do this automatically.

%% file: _approach_topp.tex
Although \cite{Pham18tro_NewApproachTimeOptimal} solves the LP in $\mathcal{O}(N)$ time using a reachability analysis approach, we can instead use factor graph variable elimination to derive an equivalent algorithm.
We hope to be clear enough that factor graph elimination is intuitive to readers, but we provide a brief introduction in the \hyperref[sec:app:factor_graphs]{Appendix} for those who feel more comfortable with one.

%%%%%%%%%%%%%%%%%%%%%%%%%%%%%%%%%%%%%%%%%%%%%%%%%%%%%%%%%%%%%%%%%%%%%%%%%%%%%%%%
% Moved to appendix
% \input{_appendix_fg_background}
%%%%%%%%%%%%%%%%%%%%%%%%%%%%%%%%%%%%%%%%%%%%%%%%%%%%%%%%%%%%%%%%%%%%%%%%%%%%%%%%

\begin{figure}
  \centering
  \input{figs/TOPP_0.tikz}
  % \footnotesize
  % \begin{maxi!}|l|[1]
  %   {\substack{x_0, \ldots, x_N,\\u_0, \ldots, u_{N-1}}}{\sum_{i=0}^N w^k x_k \label{eq:topp_objective_final}}{\label{eq:topp_problem_final}}{}
  %   \addConstraint{\bm{a}_ku_k + \bm{b}_kx_k + \bm{c}_k}{\in \mathscr{C}_k \label{eq:topp_problem_constraint_discr}}{,\quad i=0,\ldots,N}
  %   \addConstraint{x_{i+1}-x_k-2u_k\deltas}{= 0 \label{eq:topp_problem_dynamics}}{,\quad i=0,\ldots, N-1}
  %   \addConstraint{x_k}{> 0}{,\quad i=0,\ldots, N. \label{eq:topp_problem_x_positive}}
  % \end{maxi!}
  \caption{Factor graph graphically representing a 4-timestep instance of the TOPP problem \eqref{eq:topp_problem_final}.  Each variable node represents a variable $x_k$ or $u_k$ in the LP.  Each factor node represents a constraint (square) or objective (dot).
  }
  \label{fig:topp_factor_graph}
% \end{figure}
\vspace*{1.45\baselineskip}
% \begin{figure}
  \centering
  \input{figs/TOPP_u0.tikz}
  \footnotesize
  \begin{maxi*}|l|
    {\substack{x_0, \ldots, x_N,\\u_0, \ldots, u_{N-1}}}{
      % \sum_{i=0}^N w^k x_k ~~\leftarrow
      \eqref{eq:topp_objective_final}}{}{}
    \addConstraint{\eqref{eq:new_factor_after_u0}, \eqref{eq:topp_problem_constraint_discr}, \eqref{eq:topp_problem_dynamics}, \eqref{eq:topp_problem_x_positive}}{}{}
    % \addConstraint{\bm{a}_0\left( \frac{x_1 - x_0}{2\deltas} \right) + \bm{b}_0x_0 + \bm{c}_0}{\in \mathscr{C}_0 }
    % \addConstraint{\bm{a}_ku_k + \bm{b}_kx_k + \bm{c}_k}{\in \mathscr{C}_k }{,\quad i=1,\ldots,N}
    % \addConstraint{x_{i+1}-x_k-2u_k\deltas}{= 0 }{,\quad i=1,\ldots, N-1}
    % \addConstraint{x_k}{> 0}{,\quad i=0,\ldots, N. }
  \end{maxi*}\vspace*{-2\baselineskip}
  \normalsize
  \caption{Factor graph (top) and the equivalent optimization problem (bottom) after eliminating $u_0$.  The arrows denote a \emph{conditional} $u_0^*(x_0, x_1)$ and the new factor \eqref{eq:new_factor_after_u0} is the result of substituting $u_0^*(x_0, x_1)$ into \eqref{eq:topp_problem_constraint_discr}.}
  \label{fig:topp_factor_graph_u0}
% \end{figure}
\vspace*{1.45\baselineskip}

% \begin{figure}
  \centering
  \input{figs/TOPP_x0.tikz}
  \caption{Factor graph after eliminating $u_0, x_0$.  The arrow \eqref{eq:x0_solution} represents a 1-d LP we will backsubstitute at the end, and the new factor \eqref{eq:topp_new_factor_on_x1} denotes the scalar inequality constraint propagated to $x_1$ after eliminating $x_0$.}
  \label{fig:topp_factor_graph_1}
% \end{figure}
\vspace*{1.45\baselineskip}

\newcommand\groupequation[2][17pt]{%
  \setbox0=\hbox{$\displaystyle#2$}%
  \stackengine{0pt}{\copy0}{%
    \makebox[\linewidth]{\hfill$\left.\rule{0pt}{\ht0}\right\}$\kern#1}}
    {O}{c}{F}{T}{L}
}

% \begin{figure}
  \centering
  \input{figs/TOPP_bayes.tikz}
  \footnotesize
  \begin{equation}
    u_k^*(x_k, x_{k+1}) = \frac{1}{2\deltas}(x_{k+1} - x_k).\label{eq:topp_u_solution}
  \end{equation}
  \begin{maxi}|s|
    {x_k}{x_k}{\label{eq:topp_x_sol}}{\hspace*{-0.2em} x_k^*(x_{k+1}) =}
    % \rdelim\}{3}{3mm}
    \addConstraint{\frac{\bm{a}_k}{2\deltas}(x_{k+1} - x_k) + \bm{b}_kx_k + \bm{c}_k}{\in \mathscr{C}_k}{}
    \addConstraint{x_{k,min} \le x_k \le \hspace*{1.3em}}{\hspace*{-0.9em} x_{k, max}}{}
    \addConstraint{x_k}{> 0.}{}
    \raisebox{+1.8\baselineskip}[0pt][0pt]{$
      \left.\kern-\nulldelimiterspace
      \begin{array}{ @{} c } \mathstrut \\ \mathstrut \\ \mathstrut \\ \mathstrut \\ \mathstrut \\ \mathstrut \end{array}
      \right\}
      $}
  \end{maxi}
  \normalsize
  \caption{After eliminating all the variables, we obtain a \emph{Bayes Net}.  Arrows denote conditionals \eqref{eq:topp_u_solution}\eqref{eq:topp_x_sol} which we can efficiently back-substitute.}
  \label{fig:topp_bayes_net}
\end{figure}

\subsubsection*{The Factor Graph for TOPP}
The factor graph for this problem is given in Fig. \ref{fig:topp_factor_graph}.
We will proceed eliminating one variable at a time in the order $u_0, x_0, u_1, x_1, \ldots, x_{N-1}, x_N$.

\subsubsection*{Eliminating \texorpdfstring{$u_0$}{u0}}
Following the reachable set elimination ordering, we first eliminate $u_0$ by solving the LP derived by collecting only the terms in \eqref{eq:topp_problem_final} that contain $u_0$:
\begin{maxi*}|l|
  {u_0}{\mathrm{(nothing)}}{}{u_0^*(x_0, x_1) =}
  \addConstraint{\bm{a}_0u_0 + \bm{b}_0x_0 + \bm{c}_0}{\in \mathscr{C}_0}{}
  \addConstraint{x_1-x_0-2u_0\deltas}{= 0}{}
\end{maxi*}

In this case, although we have no objective function, the solution is obvious because the dynamics fully constrain $u_0$:
\begin{equation}
  u_0^*(x_0, x_1) = \frac{1}{2\deltas}(x_1 - x_0).\label{eq:topp_u0_solution}
\end{equation}
This optimal assignment is called a \emph{conditional} (this would be $p(u_0|x_0, x_1)$ in PGM literature) and is denoted by arrows in Fig. \ref{fig:topp_factor_graph_u0}.
We then substitute $u_0^*(x_0, x_1)$ into \eqref{eq:topp_problem_constraint_discr} to create a new factor on the separator $\mathcal{S}(u_0) = \{x_0, x_1\}$:
\begin{align}
  % \bm{a}_0\left( \frac{x_1 - x_0}{2\deltas} \right) + \bm{b}_0x_0 + \bm{c}_0 &\in \mathscr{C}_0.
  \bm{a}_0\left( (x_1 - x_0) / (2\deltas) \right) + \bm{b}_0x_0 + \bm{c}_0 &\in \mathscr{C}_0.
  % \nonumber \\[1em]
  % \frac{\bm{a}_0}{2\deltas}x_1 + \left(\bm{b}_0 - \frac{\bm{a}_0}{2\deltas}\right)x_0 + \bm{c}_0 &\in \mathscr{C}_0
  \label{eq:new_factor_after_u0}
\end{align}
After eliminating $u_0$, our factor graph looks like Fig. \ref{fig:topp_factor_graph_u0}.

\subsubsection*{Eliminating \texorpdfstring{$x_0$}{x0}}
Next, eliminate $x_0$ much the same way:
\begin{maxi!}|s|
  {x_0}{x_0}{\label{eq:x0_solution}}{\hspace*{-0.2em} x_0^*(x_1) =}
  \addConstraint{\frac{\bm{a}_0}{2\deltas}(x_1 - x_0) + \bm{b}_0x_0 + \bm{c}_0}{\in \mathscr{C}_0}{\label{eq:const_on_x0_x1}}
  \addConstraint{x_0}{> 0.}{\label{eq:x0_positive_after_u0}}
\end{maxi!}

For the purposes of variable elimination, we don't need to symbolically solve this LP, but instead we just need 2 things:
\begin{enumerate}
  \item conditional: $x_0^*(x_1)$, and
  \item new factor: the resulting objectives/constraints on $x_1$ after we substitute $x_0=x_0^*(x_1)$.
\end{enumerate}

For 1, we do not need an analytical expression (yet) so we just store the conditional as the optimization problem \eqref{eq:x0_solution}.
% However, we never actually need to compute 1 until the very end when we return the solution, so we delay computation until the backward pass at the end.

For 2, the new factor will consist of an objective component and a constraint component.

The objective component is easy: we can ignore it because we will select $x_1$ greedily.  More formally, our new objective factor will be $x_0^*(x_1)$, but because we also have a pre-existing factor $wx_1$ where $w$ is a very large number, our new objective term is negligible in comparison ($x_0^*(x_1) \ll wx_1$).

The constraint component of our new factor can be solved with 2 LPs the same way as in \cite{Pham18tro_NewApproachTimeOptimal}.  Since $x_1$ is a scalar, the resulting constraint on $x_1$ will take the form:
\begin{align}
  x_{1, min} \le x_1 \le x_{1, max}. \label{eq:topp_new_factor_on_x1}
\end{align}
We compute the smallest and largest possible values of $x_1$ that satisfy \eqref{eq:const_on_x0_x1}, \eqref{eq:x0_positive_after_u0}:
\begin{align}
x_{1, min} &= \underset{x_0, x_1}{\mathrm{minimize}} ~~ x_1 ~~ \mathrm{subject~to} ~ \eqref{eq:const_on_x0_x1}, \eqref{eq:x0_positive_after_u0}, \\
x_{1, max} &= \underset{x_0, x_1}{\mathrm{maximize}} ~~ x_1 ~~ \mathrm{subject~to} ~ \eqref{eq:const_on_x0_x1}, \eqref{eq:x0_positive_after_u0}. \label{eq:x1_max}
\end{align}
These are very easy to solve in just a few dozen lines of code since we need only optimize over 2, scalar variables.

% \begin{mini*}|l|
%   {x_0, x_1}{x_1}{}{x_{1, min} =}
%   % \addConstraint{\frac{\bm{a}_0}{2\deltas}(x_1 - x_0) + \bm{b}_0x_0 + \bm{c}_0}{\in \mathscr{C}_0}{}
%   \addConstraint{\eqref{eq:const_on_x0_x1}, \eqref{eq:x0_positive_after_u0}}{}
% \end{mini*}
% \begin{maxi*}|l|
%   {x_0, x_1}{x_1}{}{x_{1, max} =}
%   \addConstraint{\frac{\bm{a}_0}{2\deltas}(x_1 - x_0) + \bm{b}_0x_0 + \bm{c}_0}{\in \mathscr{C}_0}{}
%   \addConstraint{x_0}{> 0.}{}
% \end{maxi*}

% Note when solving these that \eqref{eq:const_on_x0_x1} can be projected into $\mathbb{R}^2$ by taking the planar slice of $\mathscr{C}_0-\bm{c}_0$ in the plane formed by $\bm{a}_0$ and $\bm{b}_0$.

% Then the new factor's constraint component is:
% \begin{align}
%   x_{1, min} \le x_1 \le x_{1, max}.
% \end{align}

After eliminating $u_0, x_0$ we have Fig. \ref{fig:topp_factor_graph_1}.
We repeat the elimination process on $u_1, x_1,\ldots,x_{N-1}, x_N$ until all variables are eliminated.  The result is the \emph{Bayes Net} in Fig. \ref{fig:topp_bayes_net}.

\subsubsection*{Back-substitution}
We solve the Bayes Net by back-substitution.  The final elimination step will produce a \emph{marginal} on $x_N$ ($p(x_N)$ in PGM literature) of the form:
\begin{maxi}|s|
  {x_N}{x_N}{}{x_N^* =}
  \addConstraint{x_{N,min} \le x_N \le x_{N,max}}{}{}
  \addConstraint{x_N > 0}{}{}
\end{maxi}
% \max ~~x_N~~ \mathrm{s.t.}~ x_{N,min} \le x_N \le x_{N,max}, x_N>0$.
whose solution is clearly $x_N^*=x_{N,max}$.

Then, we can compute $x_{N-1}^*=x_{N-1}^*(x_N)$ by substituting $x_N\leftarrow x_N^*$ into the conditional \eqref{eq:topp_x_sol} and solving the now single-variable scalar LP (which is just iterating through the inequalities to find the lower bound) for $x_{N-1}$.  This process is repeated until all variables are evaluated, and the resulting sequence $x_0^*, \ldots, x_N^*, u_0^*, \ldots, u_{N-1}^*$ is the solution to \eqref{eq:topp_problem_final}.

% \subsubsection*{Final Solution}

% The optimal time parameterization $s^*(t)$ can be obtained by integration:
% \begin{align}
%   s^*(t) = \int_0^t 1/\dot{s}^*(t) dt. 
% \end{align}
Finally, the optimal time parameterization $s^*(t)$ can be obtained by integrating $x^*=\dot{s}^*$.  We defer to \cite{Pham18tro_NewApproachTimeOptimal} for the intracacies of parameterizing solution.  As in \cite{Pham18tro_NewApproachTimeOptimal}, zero-inertia points are accurate in the limit $\deltas\rightarrow0$.
% $s^*(t)=\int_0^t\int_0^{\tau'}\ddot{s}^*(\tau)d\tau d\tau'$, % = \int_0^t \sqrt{x^*(\tau)}d\tau$,
% linearly interpolating $\dot{s}(t)$ from $\sqrt{x_k}$ because $u(t)=\ddot{s}(t)$ is piecewise constant. % so $\dot{s}(t)=\sqrt{x(t)}$ is piecewise linear.
% double integrating $\ddot{s}^*(t) = u^*(t(s))$ or by single-integrating $x^*(t) = \dot{s}^*(t)^2$.  However, since we made a zero-order hold assumption on $u_i$, then $u(t)$ will be piece-wise constant which makes double-integrating straightforward.  In contrast, $x$ is piecewise linear in $s$ rather than $t$ which makes integration marginally more tricky.  For each $i$, we double-integrate the constant $u_i$ (result: quadratic) until we reach the next $s_{i+1}$ (we can analytically determine the time $t_{i+1}$ associated with $s_{i+1}$ as the time the quadratic reaches the value $s_{i+1}$).  Stitching together all the segments gives the piecewise-quadratic time optimal path parameterization $s^*(t)$.
The time optimal trajectory is $\vecq^*(t) = \vecq(s^*(t))$.

%% file: figs/TOPP_0.tikz
\begin{tikzpicture}

  % Define variables
  \foreach \i in {0,...,3} {
      \node[draw, circle, minimum size=2em] (x\i) at (\i*2, 0) {$x_\i$};
  }
  \foreach \i in {0,...,2} {
      \node[draw, circle, minimum size=2em] (u\i) at (\i*2+1, -1.732) {$u_\i$};
  }
  
  % Define factors
  \foreach \i in {0,...,3} {
    \node[fill, circle, inner sep=1.5pt, label=above:{\scriptsize\eqref{eq:topp_objective_final}}] (objective\i) at (\i*2-0.3, 0.75) {};
    \node[fill, minimum size=4pt, inner sep=1.5pt, label=above:{\scriptsize\eqref{eq:topp_problem_x_positive}}] (positive\i) at (\i*2+0.3, 0.75) {};
  }
  \foreach \i in {0,...,2} {
    \node[fill, minimum size=4pt, inner sep=1.5pt, label=above:{\scriptsize\eqref{eq:topp_problem_dynamics}}] (dynamics\i) at (\i*2+1, -1.732 + 2/1.732) {};
      \node[fill, minimum size=4pt, inner sep=1.5pt, label=below:{\scriptsize\eqref{eq:topp_problem_constraint_discr}}] (constraints\i) at (\i*2+0, -2/1.732) {};
  }
  
  % Connect variables and factors with edges
  \foreach \i [evaluate=\i as \j using int(\i+1)] in {0,...,3} {
    \draw (x\i) -- (objective\i);
    \draw (x\i) -- (positive\i);
  }
  \foreach \i [evaluate=\i as \j using int(\i+1)] in {0,...,2} {
    \draw (x\i) -- (constraints\i) -- (u\i);
    \draw (x\i) -- (dynamics\i) -- (u\i);
    \draw (dynamics\i) -- (x\j);
  }
  
\end{tikzpicture}

%% file: figs/TOPP_u0.tikz
\begin{tikzpicture}

  % Define variables
  \foreach \i in {0,...,3} {
    \node[draw, circle, minimum size=2em] (x\i) at (\i*2, 0) {$x_\i$};
  }
  \foreach \i in {0,...,2} {
    \node[draw, circle, minimum size=2em] (u\i) at (\i*2+1, -1.732) {$u_\i$};
  }
  
  % Define factors
  \foreach \i in {1,...,3} {
    \node[fill, circle, inner sep=1.5pt, label=above:{\scriptsize\eqref{eq:topp_objective_final}}] (objective\i) at (\i*2-0.3, 0.75) {};
    \node[fill, minimum size=4pt, inner sep=1.5pt, label=above:{\scriptsize\eqref{eq:topp_problem_x_positive}}] (positive\i) at (\i*2+0.3, 0.75) {};
  }
  \foreach \i in {1,...,2} {
    \node[fill, minimum size=4pt, inner sep=1.5pt, label=above:{\scriptsize\eqref{eq:topp_problem_dynamics}}] (dynamics\i) at (\i*2+1, -1.732 + 2/1.732) {};
    \node[fill, minimum size=4pt, inner sep=1.5pt, label=below:{\scriptsize\eqref{eq:topp_problem_constraint_discr}}] (constraints\i) at (\i*2+0, -2/1.732) {};
  }

  \node[fill, circle, inner sep=1.5pt, label=above:{\scriptsize\eqref{eq:topp_objective_final}}] (objective0) at (0*2-0.3, 0.75) {};
  \node[fill, minimum size=4pt, inner sep=1.5pt, label=above:{\scriptsize\eqref{eq:topp_problem_x_positive}}] (positive0) at (0*2+0.3, 0.75) {};

  \node[fill, minimum size=4pt, inner sep=1.5pt, label=above:{\scriptsize\eqref{eq:new_factor_after_u0}}] (dynamics0) at (0*2+1, 0) {};

  % Connect variables and factors with edges
  \foreach \i [evaluate=\i as \j using int(\i+1)] in {0,...,3} {
    \draw (x\i) -- (objective\i);
    \draw (x\i) -- (positive\i);
  }
  \foreach \i [evaluate=\i as \j using int(\i+1)] in {1,...,2} {
    \draw (x\i) -- (constraints\i) -- (u\i);
    \draw (x\i) -- (dynamics\i) -- (u\i);
    \draw (dynamics\i) -- (x\j);
  }
  \draw (x0) -- (dynamics0) -- (x1);
  \draw[-Stealth] (x0) -- (u0) node[midway, xshift=1.1ex, yshift=1.1ex] {\scriptsize \eqref{eq:topp_u0_solution}};
  \draw[-Stealth] (x1) -- (u0) node[midway, xshift=-1.1ex, yshift=1.1ex] {\scriptsize \eqref{eq:topp_u0_solution}};
  
\end{tikzpicture}

%% file: figs/TOPP_x0.tikz
\begin{tikzpicture}
  
  % Define variables
  \foreach \i in {0,...,3} {
    \node[draw, circle, minimum size=2em] (x\i) at (\i*2, 0) {$x_\i$};
  }
  \foreach \i in {0,...,2} {
    \node[draw, circle, minimum size=2em] (u\i) at (\i*2+1, -1.732) {$u_\i$};
  }
  
  % Define factors
  % 1
  \node[fill, circle, inner sep=1.5pt, label=above:{\scriptsize\eqref{eq:topp_objective_final}}] (objective1) at (1*2-.2, 0.75) {};
  \node[fill, minimum size=4pt, inner sep=1.5pt, label=above:{\scriptsize\eqref{eq:topp_problem_x_positive}}] (positive1) at (1*2+0.3, 0.75) {};
  \node[fill, minimum size=4pt, inner sep=1.5pt, label=above:{\scriptsize\eqref{eq:topp_new_factor_on_x1}}] (new) at (1*2-0.7, 0.75) {};
  % 2, 3
  \foreach \i in {2,...,3} {
    \node[fill, circle, inner sep=1.5pt, label=above:{\scriptsize\eqref{eq:topp_objective_final}}] (objective\i) at (\i*2-0.3, 0.75) {};
    \node[fill, minimum size=4pt, inner sep=1.5pt, label=above:{\scriptsize\eqref{eq:topp_problem_x_positive}}] (positive\i) at (\i*2+0.3, 0.75) {};
  }
  \foreach \i in {1,...,2} {
    \node[fill, minimum size=4pt, inner sep=1.5pt, label=above:{\scriptsize\eqref{eq:topp_problem_dynamics}}] (dynamics\i) at (\i*2+1, -1.732 + 2/1.732) {};
    \node[fill, minimum size=4pt, inner sep=1.5pt, label=below:{\scriptsize\eqref{eq:topp_problem_constraint_discr}}] (constraints\i) at (\i*2+0, -2/1.732) {};
  }
  
  % Connect variables and factors with edges
  \draw[-Stealth] (x0) -- (u0) node[midway, xshift=1.1ex, yshift=1.1ex] {\scriptsize \eqref{eq:topp_u0_solution}};
  \draw[-Stealth] (x1) -- (u0) node[midway, xshift=-1.1ex, yshift=1.1ex] {\scriptsize \eqref{eq:topp_u0_solution}};
  \draw[-Stealth] (x1) -- (x0) node[midway, above] {\scriptsize \eqref{eq:x0_solution}};
  \draw (x1) -- (new);
  \foreach \i [evaluate=\i as \j using int(\i+1)] in {1,...,3} {
    \draw (x\i) -- (objective\i);
    \draw (x\i) -- (positive\i);
  }
  \foreach \i [evaluate=\i as \j using int(\i+1)] in {1,...,2} {
    \draw (x\i) -- (constraints\i) -- (u\i);
    \draw (x\i) -- (dynamics\i) -- (u\i);
    \draw (dynamics\i) -- (x\j);
  }
  
\end{tikzpicture}

%% file: figs/TOPP_bayes.tikz
\begin{tikzpicture}

  % Define variables
  \foreach \i in {0,...,3} {
      \node[draw, circle, minimum size=2em] (x\i) at (\i*2, 0) {$x_\i$};
  }
  \foreach \i in {0,...,2} {
      \node[draw, circle, minimum size=2em] (u\i) at (\i*2+1, -1.732) {$u_\i$};
  }
  
  % Define factors

  % Connect variables and factors with edges
  \foreach \i [evaluate=\i as \j using int(\i+1)] in {0,...,2} {
    \draw[-Stealth] (x\i) -- (u\i) node[midway, xshift=1.1ex, yshift=1.1ex] {\scriptsize \eqref{eq:topp_u_solution}};
    \draw[-Stealth] (x\j) -- (u\i) node[midway, xshift=-1.1ex, yshift=1.1ex] {\scriptsize \eqref{eq:topp_u_solution}};
    \draw[-Stealth] (x\j) -- (x\i) node[midway, above] {\scriptsize \eqref{eq:topp_x_sol}};
  }
  
\end{tikzpicture}

%% file: _approach_quadratic.tex
The variable elimination algorithm naturally extends to other objectives because it remains unchanged no matter the objectives or constraints; only the algebra of each elimination step changes.
% This way, the factor graph interpretation naturally allows us to extend the TOPP-RA algorithm to solve the trajectory retiming problem with quadratic or any other objectives for which we could perform single scalar variable elimination analytically.
Let us then define our (discretized) general quadratic objective problem as:
\begin{argmini!}|l|
  {\substack{x_0, \ldots, x_N,\\u_0, \ldots, u_{N-1}}}
    {
      % \xt_N Q_N \xt_N + 
      \sum_{i=0}^N \xt_kQ_k\xt_k + \ut_kR_k\ut_k + \xt_kN_k\ut_k\label{eq:quadratic_objective}}
    % {\norm{\xt_N}_{Q_N} + \sum_{i=0}^N \norm{\xt_k}_{Q_k} + \norm{\ut_k}_{R_k} + \norm{\ut_k}_{N_k}\label{eq:quadratic_objective}}
  {\label{eq:quadratic_problem}}{}
  \addConstraint{\eqref{eq:topp_problem_constraint_discr}, \eqref{eq:topp_problem_dynamics}, \eqref{eq:topp_problem_x_positive}}{}{}
  % \addConstraint{\bm{a}_ku_k + \bm{b}_kx_k + \bm{c}_k}{\in \mathscr{C}_k \label{eq:quadratic_problem_constraint_discr}}{,\quad i=0,\ldots,N}
  % \addConstraint{x_{i+1}-x_k-2u_k\deltas}{= 0 \label{eq:quadratic_problem_dynamics}}{,\quad i=0,\ldots, N-1}
  % \addConstraint{x_k}{> 0}{,\quad i=0,\ldots, N. \label{eq:quadratic_problem_x_positive}}
\end{argmini!}
where \emph{scalars} $\xt_k := x_k - x_{k,desired}$, $\ut_k := u_k - u_{k,desired}$, and $Q_k, R_k, N_k$ are state, control, and cross cost weights.
% \begin{argmini!}|l|
%   {\substack{x_0, \ldots, x_N,\\u_0, \ldots, u_{N-1}}}
%   {\sum_{i=0}^N \raisebox{-0.6em}{$\begin{aligned}
%     \coeff{1}{k} x_k^2 &+ \coeff{2}{k} x_ku_k + \coeff{3}{k} u_k^2 \\&+ \coeff{4}{k} x_k + \coeff{5}{k} u_k + \coeff{6}{k}
%   \end{aligned}$}\label{eq:quadratic_objective}}
%   {\label{eq:quadratic_problem}}{}
%   \\[-.5em]\nonumber
%   \addConstraint{\eqref{eq:topp_problem_constraint_discr}, \eqref{eq:topp_problem_dynamics}, \eqref{eq:topp_problem_x_positive}}{}{}
% \end{argmini!}
% where $\coeff{1}{k},\ldots,\coeff{6}{k}$ are scalar coefficients.

Elimination of $u_k$ is identical to the TOPP case; see \eqref{eq:topp_u_solution}.
Similarly, the inequality constraint portion of the new factor when eliminating $x_k$ is identical to the TOPP case; see \eqref{eq:topp_new_factor_on_x1}--\eqref{eq:x1_max}.
Then, the new objective portion of the new factor when eliminating $x_k$ is the only portion that changes (which requires analytically computing the conditional as well).

\vspace*{0.5ex} % Not sure why the spacing is glitching-out here
% Now, the only additional work required to extend TOPP-RA to quadratic objectives is to compute the conditional and new factor when we eliminate an $\xt_k$ variable.
Let us begin by computing the conditional $\xt_0^*(\xt_1)$:
{\small
\begin{argmini!}|s|
  {\xt_0}
    {\xt_0Q_0'\xt_0 + \xt_1R_0'\xt_1 + \xt_0N_0'\xt_1 \label{eq:quad_x0_objective}}
  {\label{eq:quad_x0_solution}}{\hspace*{-0.2em} \xt_0^*(\xt_1) =}
  \addConstraint{\frac{\bm{a}_0}{2\deltas}(x_1 - x_0) + \bm{b}_0x_0 + \bm{c}_0}{\in \mathscr{C}_0}{\label{eq:quad_const_on_x0_x1}}
  \addConstraint{x_0}{> 0.}{\label{eq:quad_x0_positive_after_u0}}
\end{argmini!}}%
% {\small
% \begin{argmini!}|s|
%   {x_0}
%     {\coeff{1}{0}'x_0^2+ \label{eq:quad_x0_objective}}
%   {\label{eq:quad_x0_solution}}{\hspace*{-0.2em} \xt_0^*(\xt_1) =}
%   \addConstraint{\frac{\bm{a}_0}{2\deltas}(x_1 - x_0) + \bm{b}_0x_0 + \bm{c}_0}{\in \mathscr{C}_0}{\label{eq:quad_const_on_x0_x1}}
%   \addConstraint{x_0}{> 0.}{\label{eq:quad_x0_positive_after_u0}}
% \end{argmini!}}%
where $Q_0',R_0',N_0'$ were derived after eliminating $\ut_0$ by substituting $\ut_0^* = \frac{1}{2\deltas}(x_1-x_0)$:
% Substituting $\ut_0^* = \frac{1}{2\deltas}(x_1-x_0)$, the objective function is:
% \begin{equation}
%   \underset{x_0}{\mathrm{min}} \quad \xt_0Q_0'\xt_0 + \xt_1R_0'\xt_1 + \xt_0N_0'\xt_1 \label{eq:quad_x0_solution2}
% \end{equation}\vspace*{-1\baselineskip}
{\small
\begin{align}
  Q_0' &:= Q_0 + \frac{1}{4\deltas^2}R_0 - \frac{1}{2\deltas}N_0 \\
  R_0' &:= \frac{1}{4\deltas^2}R_0 \\
  N_0' &:= -\frac{1}{2\deltas^2}R_0 + \frac{1}{2\deltas}N_0.
\end{align}}
In contrast to the TOPP case where we could use the bang-bang property to delay solving of the LP until backsubstitution,
% Whereas in the TOPP case we only needed to propagate the inequality to the new factor thanks to the greedy bang-bang property,
% we must now propagate the new objective as well.
% As a result,
we must actually solve this QP symbolically as a function of $x_1$.  Fortunately, this is tractable for a 2-dimensional (2 scalar variables) QP.

The solution to \eqref{eq:quad_x0_solution} is well-known to be piecewise linear.  As an intuition, without inequality constraints this is just the minimum of a scalar quadratic function:
\begin{align}
  \xt_{0,unconstr}^*(\xt_1) &= -b/(2a) \\
  &= -\frac{N_0'}{2Q_0'}\xt_1.
\end{align}
After inclusion of inequality constraints, the solution will be unchanged when no constraints are violated, but be the convex feasible boundary when any are violated.

% Denoting $p(\cdot)$ to be a piecewise linear function, we can express the conditional as:
% \begin{equation}
%   \xt_0^*(\xt_1) &:= p_0^*(\xt_1).
% \end{equation}
% The new objective is the piecewise quadratic obtained by substituting the piecewise $\xt_0^*(\xt_1)$ into \eqref{eq:quad_x0_solution}:
% \begin{align}
%   p_1(\xt_1)^TQp_1(\xt_1)
% \end{align}

We then substitute the piecewise linear $\xt_0^*(\xt_1)$ into \eqref{eq:quad_x0_objective} to obtain a piecewise quadratic new objective on $\xt_1$.  When eliminating $\xt_1$, the problem will have a similar form as \eqref{eq:quad_x0_solution} but with a piecewise quadratic in place of \eqref{eq:quad_x0_objective}.  Nevertheless, $\xt_1^*(\xt_2)$ will still result be piecewise linear solution as is a well-known result in parametric QP literature.

While the computational complexity of variable elimination for TOPP is exactly identical to \cite{Pham18tro_NewApproachTimeOptimal} (3 trivial LPs for each timestep), the worst-case computational complexity for quadratic objectives will be worse because the number of segments in the piecewise quadratic objectives when eliminating $\xt_k$ could be at most $\sum_{j=0}^{k-1} \mathcal{I}_j+1$ where $\mathcal{I}_j$ denotes the number of inequality constraints on $\xt_j$ in the original problem.  In practice, however, we found that the number of segments in each piecewise quadratic objective typically did not grow with problem size thus maintaining linear time complexity with respect to the number of timesteps.

%% file: _results.tex
\begin{figure}
  \centering
  \includegraphics[width=3.0in]{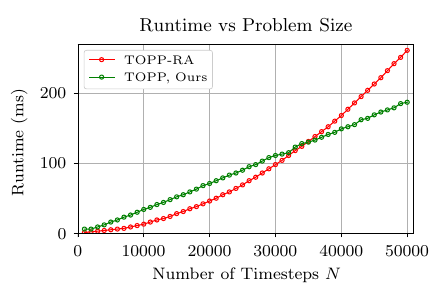}
  \caption{Computation time for a sample TOPP problem with varying problem size shows (1) our algorithm is linear-time and (2) is faster than TOPP-RA.  Both algorithms are implementationed in C++, compiled with identical compiler flags \texttt{-O3;-flto}, and run on the same M1 Macbook Air.}
  \label{fig:runtime_topp}
\end{figure}

To experimentally validate our algorithm, we implement a C++ version of our algorithm \cite{Chen23_ImplementationGeneralizingTrajectory} to analyze runtime then apply it to a cable robot to demonstrate the efficacy of quadratic objectives over the time-optimal objective.

% Our implementation is available on github\footnote{\href{https://github.com/gchenfc/gtsam/tree/features/gerry/trajectory_retiming/gtsam_unstable/retiming}{github.com/gchenfc}}.

\subsection{Runtime} \label{ssec:runtime}

To validate the runtime and complexity of our algorithm, we implement a C++ version of our algorithm and compare it to the state-of-the-art TOPP-RA algorithm \cite{Pham23_ToppraTimeOptimalPath} on a sample problem with varying problem size.

As shown in Figure \ref{fig:runtime_topp}, our implementation is linear-time and has similar speed to TOPP-RA.
We test a simple problem: $x_{k+1} = x_k + (0.5)u_k$ with constraint $x_k + u_k \le 0.1$.  The results are identical to 7 decimal places.
% It is unclear why TOPP-RA appears to have super-linear runtime.
% Although TOPP-RA appears to have a super-linear runtime, more than 1000 timesteps is rarely necessary so sub-millisecond timing is expected from both algorithms.
Although TOPP-RA appears to have a super-linear runtime, in most applications a few hundred timesteps is sufficient so sub-millisecond timing is expected from both algorithms.

\begin{figure}
  \centering
  \includegraphics[width=3.0in]{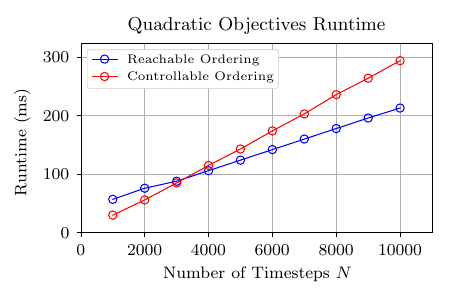}
  \caption{Runtime plot for a sample quadratic objective retiming problem shows that our algorithm is still $\mathcal{O}(n)$, even with quadratic objectives.}
  \label{fig:runtime_quadratic}
\end{figure}

\begin{figure}
  \centering
  \includegraphics[width=3.0in]{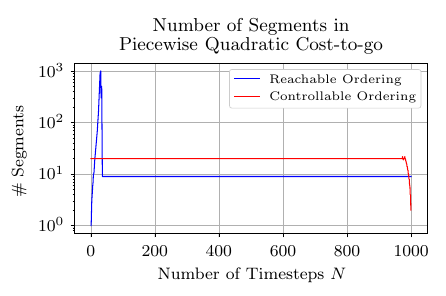}
  \caption{The number of segments in the piecewise quadratic cost-to-go functions for each time step remains mostly constant throughout the optimization.  The rapid growth followed by decay for the reachable ordering explains the apparent non-zero y-intercept in Fig. \ref{fig:runtime_quadratic}.}
  \label{fig:segments_quadratic}
\end{figure}

Figure \ref{fig:runtime_quadratic} shows that the runtime, even with quadratic objectives, remains linear with problem size.  Figure \ref{fig:segments_quadratic} further evidences this fact by showing that the number of inequalities carried forward from each time step remains roughly constant throughout the optimization.  Objectives $x_N^2 + \sum x_k^2 + u_k^2$ were added to the problem above to generate these results.  As discussed in Section \ref{ssec:quadratic_approach}, completely general quadratic objectives 
% of the form $\xt^TQ\xt + \ut^TR\ut + \xt^TN\ut$ with $\xt:=x-x_d, \ut:=u-u_d$
are supported.

\subsection{Cable-Driven Parallel Robot Application} \label{ssec:cable_robot}
% \subsection{Defining a Trajectory Retiming Problem for a Cable-Driven Parallel Robot}

Operating at control limits for prolonged periods is often undesirable for reasons such as safety, wear and tear, noise, and energy consumption.  In such cases, it is preferable to balance multiple objectives such as matching a desired speed while also minimizing control effort.

This is especially true for cable robots, which have inherently low stiffness maintained primarily through internal tension.  Approaching control limits necessarily diminishes the possible amount of internal tension causing reduced stability and increased risk of cable slackening.
In this section, we formulate the a quadratic objective retiming problem for a cable robot path tracking problem, then quantitatively and qualitatively show results running on a real robot.

\subsubsection{Formulating the Cable Robot Tracking Problem}
Due to the nature of cable-driven parallel robots, it makes most sense to define the path tracking problem in the task space rather than the joint space.  Just as serial manipulators can express task space constraints in the joint space via the Jacobian (but the other direction is more difficult), cable robots can express joint space constraints in the task space via the wrench matrix, $\boldW$: defined by $\wrench=\boldW \vect$ implying $\vecqd = \boldW^T \vecxd$, where $\wrench,\boldW,\vect,\vecxd$ denote the force on the end-effector, wrench matrix, cable tensions, and task space velocity, respectively.
Computing feasible polytopes
$\mathscr{C}(s):=\{\wrench : \exists \vect \in [\vect^-, \vect^+] \mathrm{~s.t.~} \wrench=\boldW(s)\vect\}$%
, we define then our standard dynamics and constraints in the form (note: vector $\vecx$ is distinct from scalar $x$):
% ~\\\vspace*{-2.2\baselineskip}
\begin{subequations}\label{eq:task_space_constraints}
\begin{align}
  \boldA(s)\ddot{\vecx} + \dot{\vecx}^T\boldB(s)\dot{\vecx} + \boldf(s) &\in \mathscr{C}(s) \\
  \boldA^{v}(s)\dot{\vecx} + \boldf^{v}(s) &\in \mathscr{C}^{v}(s).
\end{align}
\end{subequations}
We then apply the process in Section \ref{ssec:reparam} to convert to \eqref{eq:topp_problem_constraint_discrete}.

For objectives, we seek to match a setpoint speed using objective
$q'\norm{\vecxd^T\vecxd - v_d^2}^2$ and maximize motor torque margin using objective $r\norm{\wrench-\boldW \vect_m}^2$, where $q', r$ are weighting factors, $v_d$ is the desired speed, and $\vect_m$ is the arithmetic mean of the minimum and maximum allowable tensions, similar to \cite{Pott09ckin_ClosedformForceDistribution,Chen22iros_LocallyOptimalEstimation}.  These can be expressed as:
\begin{align}
  q\norm{x-x_d}^2 + r\norm{\boldA u + \boldB x + \boldC'}^2
\end{align}
with $q := q'\norm{\vecx'}^4$, $x_d := v_d^2/\norm{\vecx'}^2$, $\vecx'$ denotes the derivative of $\vecx$ with respect to $s$, and $\boldA, \boldB, \boldC'$ are the same coefficients as in the equations of motion except subtracting off $\boldW\vect_m$.  These can, in turn, be combined to form a single quadratic objective in the form $Q_k\xt_k^2 + R_k\ut_k^2 + N_k\xt_k\ut_k$, computed by software.  Note we still keep \eqref{eq:task_space_constraints} for safety.

\subsection{Robot behavior with QOPP vs TOPP}
We solve the cable robot path tracking problem for a 2m/s star-shaped trajectory with 1000 discretization points.  Figs. \ref{fig:tensions_compare} and \ref{fig:cable_robot} show the solution and execution result, respectively.
Tracking error measured with OptiTrack Motion Capture.

% See Video Supplemental for a video of the robot running with quadratic objectives vs minimum-time objective.

% \begin{figure}
%   \centering
%   TOPP
%   \includegraphics[width=0.4\linewidth,trim=180px 90px 340px 160px,clip]{figs/topp_long_exposure.png}
%   \includegraphics[width=0.4\linewidth,trim=180px 90px 340px 160px,clip]{figs/qopp_long_exposure.png}
%   QOPP (ours)
%   \caption{TOPP (left) and QOPP (ours, right) used for a cable robot a path following task show that TOPP generates significantly more error than QOPP due to saturating controls leaving insufficient control margin.}
% \end{figure}

% \begin{figure}
%   \centering
%   \includegraphics*[width=0.6\linewidth]{figs/robot2_cropped.JPG}
%   \caption{Running our algorithm on the robot qualitatively shows (see Video Supplemental) that using quadratic objectives produces less oscillation by keeping a larger margin from the torque limits.}
% \end{figure}

% \missingfigure[figwidth=.7\linewidth]{Figure showing umm... tracking error for the two?  Maybe just a video?}

%% file: _discussion.tex
By extending the trajectory retiming problem to a wider domain, we potentially open opportunities for sharing techniques with the broader trajectory optimization community.

One such opportunity is to generalize to arbitrary nonlinear objectives by employing an SQP strategy in which a nonlinear optimization problem is repeatedly approximated as a QP problem over a step direction \cite{Betts93jgcd_PathconstrainedTrajectoryOptimization}.  This linear-time algorithm is specific to the scalar structure of the problem, so specialized trajectory retiming solvers may outperform general solvers even for complicated objectives.

Another opportunity is to perform trajectory optimization with alternating path and retiming optimizations.  Whereas time-optimal retiming would not apply due to differing objectives, introducing quadratic objectives brings us closer to sharing the same objectives, which would enable semi-decoupled approaches.  GTDynamics also uses GTSAM and uses a time-scaling variable.  Experiments in variable ordering have suggested such an approach may be promising \cite{Xie20arxiv_gtdynamics}.

% \cite{Byravan14icra_SpacetimeFunctionalGradient} (T-CHOMP), is doing trajectory optimization but discusses retiming explicitly.  It may be possible in this formulation to e.g. do an alternating optimization or choose a special elimination ordering to enable the temporal optimization to be solved using this efficient method, and handle the "hardest" of the inequality constraints this way too.

% Also, note that gtdynamics already tries to do this by optimizing over a single time-stretching factor.  As T-CHOMP describes, this might be easier if we can retime the whole trajectory rather than use a single time-scaling factor.

% Another idea: it is plausible that there are closed-form or more efficient bisection methods of finding intersections of feasible regions and quadratic objective minima, using a switching-algorithm approach.  But I'm not quite smart enough to figure that out.

% \cite{Betts93jgcd_PathconstrainedTrajectoryOptimization} might be a good reference for nonlinear programming by sequential quadratic.

%% file: _conclusions.tex
In this work, we proposed a linear time algorithm for solving trajectory retiming problems with quadratic objectives (QOPP).
We discussed some applications of quadratic objectives, such as balancing objectives of fast speed and keeping distance from control limits for improving safety and robot wear.
We described the algorithm by first reinterpreting TOPP-RA using factor graphs, then extending the factor graph algorithm to handle quadratic objectives.
Finally, we experimentally validated our algorithm's runtime and performance improvement on a cable robot path tracking problem.

\begin{figure}[t]
  \centering
  \vspace*{-0.2em}
  \raisebox{1.8em}{\includegraphics[width=0.43\linewidth]{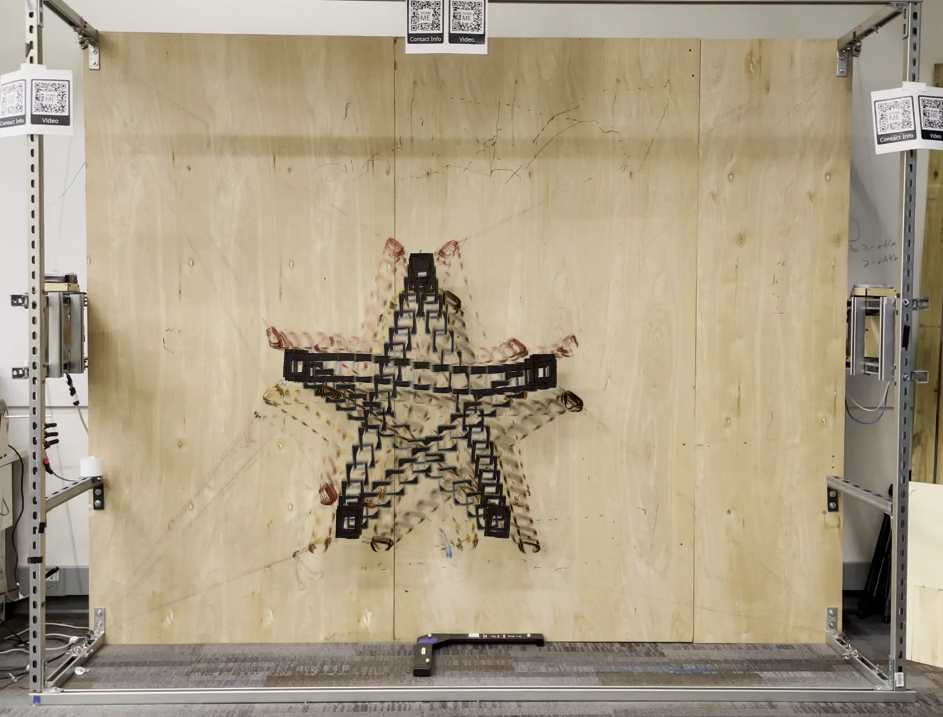}}
  \includegraphics[width=0.55\linewidth]{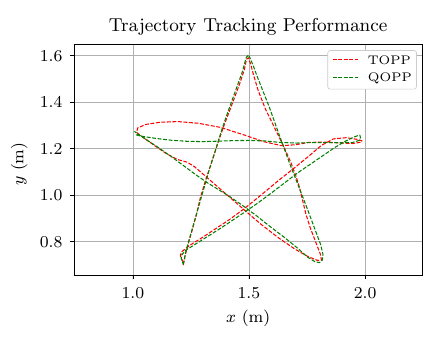}
  \caption{TOPP (red) and QOPP (ours, green) used for a cable robot path following task show that TOPP generates significantly more error (62mm max tracking error) than QOPP (42mm max tracking error) due to saturating controls.  We combine video frames to visualize the robot moving (left).}
  \label{fig:cable_robot}
\end{figure}

%% file: _appendix_fg_background.tex
% \subsubsection*{Brief Introduction to Factor Graph Variable Elimination}
\subsection*{For readers new to Factor Graph Variable Elimination:}
% For readers \emph{not} coming from a probabilistic graphical models (PGM) background,
Think of factor graph variable elimination as a graphical representation for the process of solving optimization problems ``one variable at a time''.  The optimal value for that variable is given as a function of the other variables still in the problem, and a new optimization problem is created by substituting the optimal value for that variable into the original problem.  A key detail is that only the variables that share a factor with the eliminated variable are involved in the elimination process, thereby \emph{exploiting the sparsity of the problem automatically}.  This process is repeated until only one variable remains, at which point the solution is returned.

More formally, an optimization problem can be described with a factor graph by denoting each of the variables to be optimized as a ``variable'' node and each of the optimization objective terms and constraints as a ``factor'' node, where an edge connects each variable a factor depends on (see Fig. \ref{fig:topp_factor_graph} as an example).  Then to solve the optimization problem, the variable elimination algorithm ``eliminates'' (solves) one variable, $x$, at a time, passing its constraints and objective terms as a new factor on the \emph{separator}, $\mathcal{S}(x)$: the set of variables sharing a factor with the eliminated variable.  More complete descriptions of factor graph elimination for solving optimal control problems can be found in \cite{Yang21icra_EqualityConstrainedLinear,Chen22iros_LocallyOptimalEstimation,Dellaert17fnt}.

GTSAM \cite{gtsam} is a mature C++ software library that implements factor graph variable elimination, including with quadratic objectives and linear equality constraints.  Architecturally, it allows easy extension to handle additional factor types, such as inequality constraints, which we do later in this paper.  Clickable link back to \ref{ssec:topp}.

\subsection*{For readers familiar with factor graphs (especially GTSAM):}
Consider that a factor with ``zero-covariance'' is a constraint.  Then, for example, a graph containing only Gaussian factors is equivalent to solving an equality-constrained linear least squares problem.  Please refer to \cite{Yang21icra_EqualityConstrainedLinear} for additional details, which we extend in this paper to handle inequality constraints on scalar variables for certain problem structures.